\documentclass{article} 
\usepackage{iclr2027_conference,times}

\usepackage{amsmath,amsfonts,bm}

\def\eqref#1{equation~\ref{#1}}

\def\1{\bm{1}}

\DeclareMathAlphabet{\mathsfit}{\encodingdefault}{\sfdefault}{m}{sl}
\SetMathAlphabet{\mathsfit}{bold}{\encodingdefault}{\sfdefault}{bx}{n}

\usepackage{hyperref}
\usepackage{url}
\usepackage{amsmath, amsthm, graphicx, amssymb, booktabs, xcolor}
\usepackage{float}

\title{From Scene Graphs to Answers: Selective Neuro-Symbolic Reasoning for Autonomous Driving}

\author{%
Yiyao Wang\textsuperscript{1,2}\thanks{Equal contribution.},
Pei Liu\textsuperscript{1}\footnotemark[1],
Fangzhou Liu\textsuperscript{2},
Jun Ma\textsuperscript{1}\thanks{Corresponding author.} \\
\textsuperscript{1}The Hong Kong University of Science and Technology (Guangzhou)\\
\textsuperscript{2}Harbin Institute of Technology
}

\iclrfinalcopy
\begin{document}

\maketitle

\begin{abstract}

Autonomous-driving question answering requires reasoning over structured scene information, yet existing vision-language approaches largely delegate heterogeneous reasoning operations to a single neural inference process. We argue that this uniform strategy overlooks a fundamental distinction: some queries admit exact symbolic solutions, while others require semantic interpretation. We introduce a \emph{query-adaptive neuro-symbolic reasoning framework} that explicitly allocates computation according to the nature of the query. At its core is a hierarchical Spatiotemporal Scene Graph (STSG) that separates persistent object identities from frame-specific states and represents spatial relations and temporal transitions as explicit directed structures. Given a query, a symbolic executor first attempts to resolve it through exact graph operations; only when symbolic execution abstains is an LLM invoked for semantic reasoning. For these unresolved queries, query-conditioned graph retrieval and evidence filtering preserve relation direction, temporal locality, and object semantics, providing the LLM with compact and verified task-relevant evidence. This design shifts the role of the LLM from a universal reasoning engine to a targeted semantic reasoner, while allowing deterministic computation to be handled exactly and efficiently. We evaluate the framework on 5,916 NuScenes-QA questions across all ten scenes of nuScenes v1.0-mini under an oracle-perception setting. The complete system achieves $80.63\%$ overall accuracy with GPT-5.4-mini, improving over the corresponding LLM-only configuration by 5.48 percentage points; with DeepSeek-V4-Flash, the improvement reaches 6.64 points. The largest gains occur on counting questions, with improvements of 10.20 and 12.61 points, respectively. These results demonstrate that {selective allocation of reasoning} can improve both accuracy and inference efficiency, providing a principled alternative to treating the LLM as a universal executor for structured reasoning in autonomous driving.

\end{abstract}

\section{Introduction}

Autonomous driving increasingly requires systems to move beyond perception toward structured reasoning about the surrounding world. Given a driving scene, an intelligent agent should not only recognize vehicles, pedestrians, and their attributes, but also answer relational questions such as whether an object exists in a particular region, how many objects satisfy a spatial constraint, or which object has a specific relation to the ego vehicle. Recent datasets such as nuScenes~\cite{caesar2020nuscenes} and NuScenes-QA~\cite{qian2024nuscenes} have made such reasoning increasingly accessible. However, these questions are fundamentally different from conventional visual recognition: they require identifying relevant entities, composing multiple relations, preserving relational direction, and executing exact operations such as filtering and counting. The central challenge is therefore not simply {how to make a language model reason over driving scenes}, but rather {how to allocate different reasoning operations to the computational mechanisms that are best suited for them}.

Large language models (LLMs) have emerged as powerful general-purpose reasoners and have demonstrated impressive capabilities in multi-step inference~\cite{wei2022chain,yao2022react}. In autonomous-driving QA, this naturally suggests using an LLM as a universal reasoning engine over scene information. Yet this design implicitly assumes that every query should be solved through neural language reasoning. This assumption is problematic for structured environments. Many questions have deterministic semantics: existence can be verified directly, spatial constraints can be evaluated by graph operations, and counting can be performed exactly once the relevant entities have been identified. Forcing an LLM to reproduce these operations through free-form generation introduces an unnecessary source of uncertainty and computation. Conversely, questions involving semantic ambiguity or compositional interpretation may not be naturally captured by a fixed symbolic rule system. This reveals a fundamental asymmetry: {some reasoning is computable, while other reasoning is interpretive}. Treating both uniformly as neural generation leaves this distinction unexplored.

This observation motivates a different view of LLM-based reasoning: {the key problem is not choosing between symbolic and neural reasoning, but deciding when each should be used}. Rather than using an LLM as the executor of every reasoning step, we formulate reasoning as a {query-adaptive computation process}. A query that admits an unambiguous symbolic solution should be resolved exactly and directly, whereas a query that exceeds the expressiveness of the symbolic executor should be delegated to a neural reasoner. This perspective is related to the broader neuro-symbolic paradigm~\cite{yang2025neuro,fang2024large}, but differs in emphasis: instead of tightly coupling symbolic and neural modules throughout inference, we explicitly introduce {selective reasoning} through an abstention mechanism. The symbolic component is therefore not merely an auxiliary tool for the LLM; it serves as a gate that determines whether neural reasoning is necessary at all.

To realize this principle, we introduce a hierarchical Spatiotemporal Scene Graph (STSG) that provides the structured substrate for selective reasoning. The STSG separates persistent physical entities from their frame-specific states, while explicitly encoding directed spatial relations and temporal transitions. This representation allows the system to preserve object identity across observations while supporting exact relational operations at the state level. On top of the STSG, our framework first attempts symbolic execution over the complete structured scene. Only when symbolic execution abstains does the system invoke an LLM. For these unresolved queries, we further perform query-conditioned graph retrieval and evidence filtering, restricting neural reasoning to relevant entities, relations, and frame-specific evidence. This design also complements recent retrieval-augmented approaches~\cite{lewis2020retrieval,han2024retrieval,guo2025lightrag}: instead of treating retrieval merely as a mechanism for supplying more context to an LLM, we use structured retrieval as part of a {reasoning allocation pipeline} that separates exact computation from semantic inference.

We evaluate this framework on 5,916 NuScenes-QA questions across all ten scenes of nuScenes v1.0-mini. Under the oracle-perception setting, our full system achieves 80.63\% accuracy with GPT-5.4-mini, improving over the corresponding LLM-only configuration by 5.48 percentage points. With DeepSeek-V4-Flash, the improvement reaches 6.64 percentage points. The largest improvements occur on counting questions, where symbolic execution provides gains of 10.20 and 12.61 percentage points for GPT-5.4-mini and DeepSeek-V4-Flash, respectively. Selective reasoning also reduces total inference time by bypassing neural inference for deterministically solvable queries. These results indicate that the benefit of neuro-symbolic reasoning in structured driving scenes does not necessarily come from combining two reasoning modules everywhere, but from {selectively assigning computation according to the nature of the query}.
Our contributions are summarized as follows:
\begin{itemize}
\item We formulate LLM-based driving-scene QA as a {query-adaptive reasoning problem}, arguing that deterministic and semantic reasoning should not be uniformly delegated to a neural language model.
\item We introduce a hierarchical Spatiotemporal Scene Graph that separates persistent entities from frame-specific states and explicitly represents spatial and temporal relations, enabling exact structured reasoning over driving scenes.
\item We propose a selective neuro-symbolic reasoning architecture with an abstention-based routing mechanism, allowing symbolic execution to handle deterministically solvable queries while delegating unresolved cases to an LLM.
\item We demonstrate that this selective allocation of reasoning improves both accuracy and inference efficiency across two LLM backends on NuScenes-QA, with particularly substantial gains on exact counting queries.
\end{itemize}

\begin{figure}[H]
    \centering
    \includegraphics[width=\textwidth]{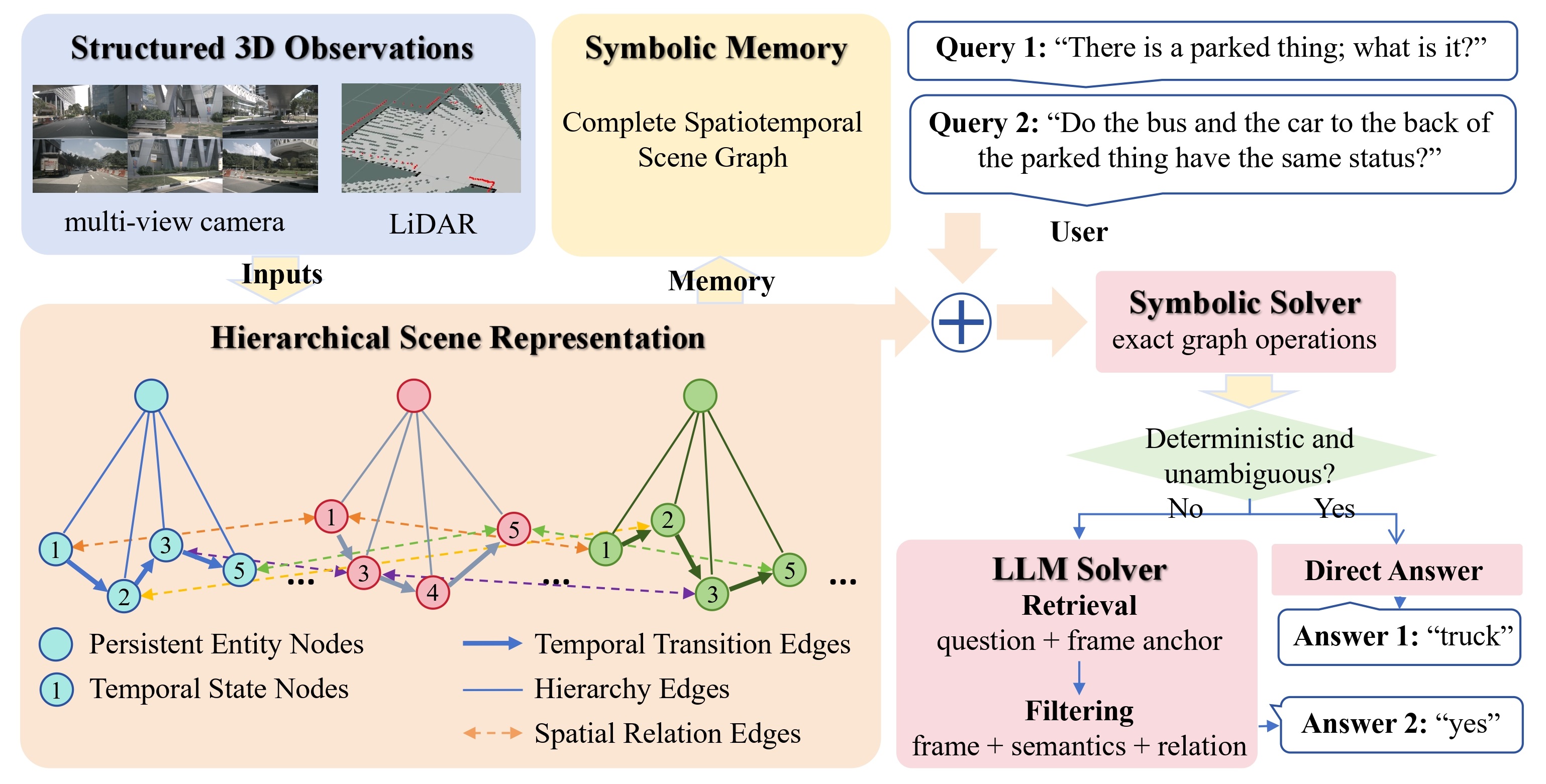}
    \vspace{-6mm} 
    \caption{\textbf{Overview of the framework.} Structured 3D observations form a hierarchical spatiotemporal scene graph that serves as memory. Unambiguous queries are answered directly by the symbolic solver; otherwise, graph retrieval and evidence filtering provide context for LLM reasoning.}
    \label{fig:pipeline}
\end{figure}

\section{Related Work}

Vision-language models connect driving-scene understanding with language-based decisions. DriveGPT4 generates driving explanations and control predictions~\cite{Xu2024DriveGPT4}. DriveVLM combines scene analysis with hierarchical planning, while its dual variant incorporates conventional perception and planning~\cite{tian2024DriveVLM}. DriveLM organizes question--answer pairs into a graph of logical dependencies~\cite{sima2025DriveLM}.

Scene graphs provide structured representations for robotic perception and reasoning. SceneGraphFusion incrementally predicts and fuses object and relation labels~\cite{wu2021SceneGraphFusion}, while Hydra constructs and optimizes hierarchical scene graphs~\cite{hughes2022Hydra}. ConceptGraphs builds open-vocabulary, object-centric maps for language-guided queries and planning~\cite{gu2023ConceptGraphs}.

Graph-based RAG exploits relational structure to organize evidence. RoG learns relation-path plans and retrieves paths for LLM reasoning~\cite{Luo2024Reasoning}. G-Retriever combines connected-subgraph retrieval with graph soft prompts and textualized evidence~\cite{He2024GRetriever}. KG$^2$RAG expands and organizes document chunks through graph connections~\cite{zhu2025KGGRAG}, while GraphRAG uses hierarchical community summaries for global questions~\cite{edge2025localglobal}. ToG-3 iteratively refines queries and evidence subgraphs through multi-agent collaboration~\cite{wu2026thinkongraph30}.

Neuro-symbolic execution and adaptive computation motivate selective reasoning. VISPROG uses an LLM to generate programs that compose pretrained modules and symbolic operations~\cite{Gupta2023Visual}. Adaptive-RAG learns to select retrieval strategies according to question complexity~\cite{jeong2024AdaptiveRAG}; RouteLLM learns model selection from preference data~\cite{ong2025RouteLLM}. Self-RAG learns reflection tokens for retrieval control and output evaluation~\cite{Asai2024SelfRAG}.

Our framework integrates symbolic-first execution with retrieval-augmented reasoning over a spatiotemporal scene graph. A conservative symbolic solver directly answers supported, unambiguous queries and invokes the LLM only upon abstention, avoiding unnecessary neural computation. This selective design requires no task-specific model fine-tuning and improves both QA accuracy and inference efficiency under the evaluated oracle-perception setting.

\section{Methodology}
\label{sec:methodology}

In this section, we present a spatiotemporal retrieval-augmented framework for multi-frame question answering in autonomous driving scenarios. The framework takes structured 3D object observations from an upstream perception module and decouples deterministic geometric reasoning from language-based inference. In our experiments, the perception interface is instantiated with ground-truth 3D annotations in order to isolate retrieval and reasoning errors from perception errors. The interface can be connected to learned 3D vision-language perception modules that produce the required observations. As illustrated in Figure~\ref{fig:pipeline}, the framework consists of three components:
\begin{enumerate}
    \item \textbf{Hierarchical Spatiotemporal Graph Construction}: organizing persistent object identities, frame-specific states, spatial relations, and temporal transitions into an explicit, multi-layered spatiotemporal graph.
    \item \textbf{Structured Graph Projection and Indexing}: projecting task-relevant graph facts into a graph-vector retrieval index while retaining the complete graph as symbolic memory;
    \item \textbf{Neuro-Symbolic Dual-Engine Reasoning}: first attempting deterministic symbolic execution and, when the symbolic solver abstains, invoking a retrieval-augmented LLM with query-conditioned evidence filtering.
\end{enumerate}

\subsection{Spatiotemporal Scene Graph Representation}
\label{subsec:scene_graph}

To represent a dynamic driving scene across multiple frames, we introduce a hierarchical graph that links persistent physical identities to their frame-specific realizations. We refer to this representation as a Spatiotemporal Scene Graph (STSG), denoted by $\mathcal{G}=(\mathcal{V},\mathcal{E})$. Unlike a collection of independent frame-level scene graphs, the proposed representation provides a shared identity layer across the sequence while preserving the state and spatial configuration of each timestamp.

The node set $\mathcal{V}$ contains persistent entity nodes $\mathcal{V}_{ent}$, temporal state nodes $\mathcal{V}_{state}$, and an ego-reference node $v_{\mathrm{ego}}$. The edge set $\mathcal{E}$ contains spatial relations $\mathcal{E}_{spat}$, entity--state hierarchy relations $\mathcal{E}_{hier}$, and temporal transitions $\mathcal{E}_{temp}$:
\begin{equation}
\mathcal{V} = \mathcal{V}_{ent} \cup \mathcal{V}_{state} \cup \{v_{\mathrm{ego}}\},
\qquad
\mathcal{E} = \mathcal{E}_{spat} \cup \mathcal{E}_{hier} \cup \mathcal{E}_{temp}.
\end{equation}

Figure~\ref{fig:Graph} in Appendix~\ref{app:gt_graph_construction} illustrates the entity--state hierarchy, temporal transitions, and spatial relations.

\subsubsection{Persistent Entity Nodes ($\mathcal{V}_{ent}$)}
Persistent entity nodes represent unique instance-level physical objects whose identities are maintained across the scene sequence. Each entity node $v_i^{ent} \in \mathcal{V}_{ent}$ provides a stable identifier that links observations of the same physical object across frames.

\subsubsection{Temporal State Nodes ($\mathcal{V}_{state}$)}
For each observation of entity $i$ at timestamp $t$, we create a temporal state node $v_{i,t}^{state}\in\mathcal{V}_{state}$. It represents the realization of that entity at the corresponding frame:
\begin{equation}
v_{i,t}^{state} = \big(i,t,\mathbf{a}_{i,t}\big),
\end{equation}
where $\mathbf{a}_{i,t}$ denotes the frame-specific semantic and physical attributes, including the object category and motion status. Separating persistent identity from temporal state prevents repeated observations of the same object from being treated as independent physical entities.

\subsubsection{Spatial Relation Edges ($\mathcal{E}_{spat}$)}
Spatial edges encode directed relations between state nodes observed at the same timestamp, as well as relations between a state node and the ego vehicle. Let $\mathbf{p}_{i,t}^{xy}$ and $\mathbf{p}_{j,t}^{xy}$ be the BEV centers of a subject object $i$ and a reference object $j$, respectively. Their spatial relations are defined as
\begin{equation}
\Delta\mathbf{p}_{i\mid j,t}
=
\mathbf{p}_{i,t}^{xy}
-
\mathbf{p}_{j,t}^{xy},
\qquad
\theta_{i\mid j,t}
=
\operatorname{atan2}
\left(
\Delta y_{i\mid j,t},
\Delta x_{i\mid j,t}
\right).
\end{equation}

Following the directional taxonomy of NuScenes-QA \cite{qian2024nuscenes}, we use its six ego-aligned spatial predicates and denote the corresponding directional mapping by $\phi(\theta)$. The resulting directed relation is
\begin{equation}
e_{i\mid j,t}^{spat} = \left(
v_{i,t}^{state}, \phi(\theta_{i\mid j,t}), v_{j,t}^{state} \right)
\in
\mathcal{E}_{spat}.
\end{equation}
When the reference is the ego vehicle, $v_{j,t}^{state}$ is replaced by $v_{\mathrm{ego}}$. This subject-reference convention makes the direction of each edge explicit and reduces the risk of reversing a spatial relation during reasoning.

\subsubsection{Hierarchy Edges ($\mathcal{E}_{hier}$)}
Hierarchy edges establish structural links between persistent global entities and their frame-specific state instances. A directed hierarchy edge $e_{i,t}^{hier} \in \mathcal{E}_{hier}$ connects a global entity $v_i^{ent}$ to its temporal state $v_{i,t}^{state}$:
\begin{equation}
e_{i,t}^{hier} = \big( v_i^{ent}, \text{\texttt{HAS\_STATE}}, v_{i,t}^{state} \big)
\end{equation}
The two-level representation allows an object to be referenced either as a persistent instance or as a frame-specific state.

\subsubsection{Temporal Transition Edges ($\mathcal{E}_{temp}$)}
Let $t_i^{(1)}<t_i^{(2)}<\cdots<t_i^{(K_i)}$ denote the ordered timestamps at which entity $i$ is observed. We connect each pair of successive observations by
\begin{equation}
e_{i,k}^{temp}
=
\left(
v_{i,t_i^{(k)}}^{state},
\text{\texttt{NEXT\_STATE}},
v_{i,t_i^{(k+1)}}^{state}
\right)
\in
\mathcal{E}_{temp}.
\end{equation}
These links form an explicit state trajectory for each persistent entity and provide a structural basis for reasoning about motion evolution, temporal ordering, and cross-frame comparisons.

\subsection{Structured Graph Projection and Indexing}
\label{subsec:graph_indexing}

The complete spatiotemporal graph $\mathcal{G}$ serves as the structured memory of the symbolic reasoning engine. For retrieval-augmented inference, we derive a retrieval-oriented graph view
\begin{equation}
\mathcal{G}_{R}
=
\Pi_{R}(\mathcal{G}),
\end{equation}
where $\Pi_R$ maps relevant graph elements into canonical node and relation records while preserving entity identifiers, relation directionality, and temporal references.

The resulting records are inserted into a LightRAG-based graph-vector index~\cite{guo2025lightrag}. This structured insertion bypasses the intermediate conversion of the graph into documents and avoids LLM-based re-extraction of entities and relations. In contrast to arbitrary document chunking, each indexed record retains the atomic subject--predicate--object semantics of the original graph.

\begin{figure}[H]
    \centering
    \includegraphics[width=\textwidth]{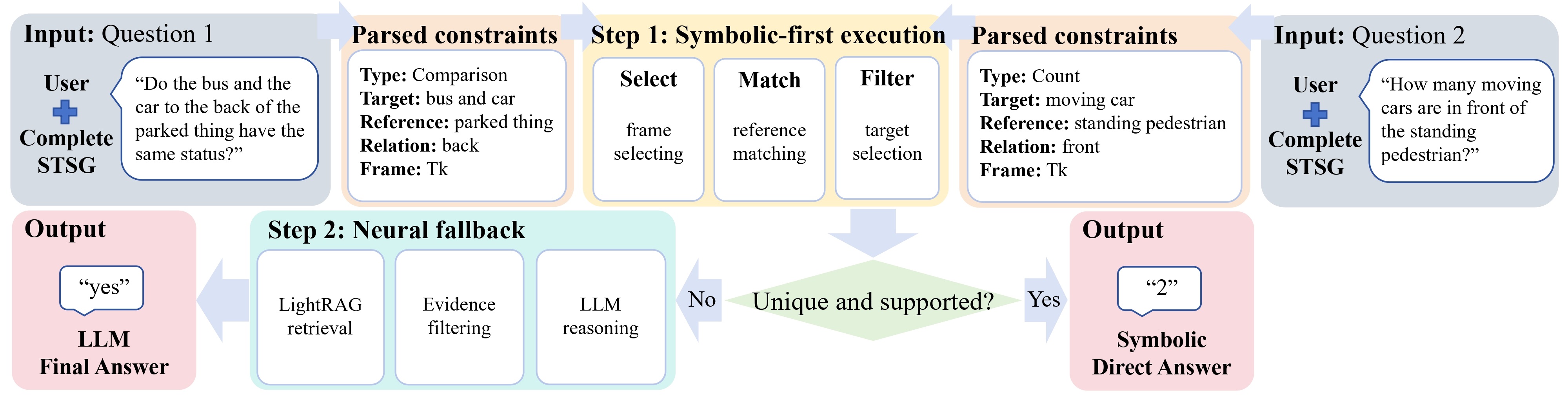}
    \vspace{-6mm} 
    \caption{\textbf{Neuro-Symbolic Question Answering.} The symbolic solver applies frame selection, reference matching, and target filtering to the complete scene graph. The counting example is answered directly as ``2'', while the unsupported comparison query triggers abstention and proceeds through LightRAG retrieval, evidence filtering, and LLM reasoning to produce ``yes''.}
    \label{fig:Answering}
\end{figure}

\subsection{Neuro-Symbolic Dual-Engine Question Answering}
\label{subsec:dual_engine}

Given a natural language query $Q$, directly providing a large graph $\mathcal{G}$ to a Large Language Model (LLM) is often impractical due to context-length and computational constraints \cite{beltagy2020longformer,liu2024lost}. Dense retrieval reduces the amount of input but may confuse relations that appear in similar contexts while differing in polarity, direction, or other semantic qualifiers, potentially leading to the retrieval of semantically incompatible facts \cite{weller2024nevir}. To address these limitations, we propose a Neuro-Symbolic Dual-Engine architecture that decouples deterministic symbolic filtering from LLM-based semantic reasoning.

\subsubsection{Conservative Symbolic-First Routing}

We define the symbolic solver as a selective solver that may abstain when a question $Q$ cannot be resolved deterministically. Let $\mathcal{Y}$ denote the task-specific answer space, including object categories, motion-status labels, counts, and binary answers. Let $\Phi_{\mathrm{sym}}$ denote the symbolic solver, $\Phi_{\mathrm{llm}}$ denote the retrieval-augmented LLM reasoning engine, and $\bot$ denote abstention.

The symbolic solver first evaluates the question against the complete spatiotemporal graph:
\begin{equation}
a_{\mathrm{sym}}
=
\Phi_{\mathrm{sym}}
\left(
Q,\mathcal{G},t_q
\right)
\in
\mathcal{Y}
\cup
\{\bot\}.
\end{equation}

For supported and unambiguous questions, the solver performs exact set operations over frame-specific state nodes. It determines existence, counts unique persistent identities for counting questions, and returns an object category or motion status when the answer is unique. When the query pattern is unsupported or multiple answers remain plausible, the solver returns $\bot$ rather than forcing a prediction.

Comparison questions and other compositionally ambiguous questions are therefore delegated to the neural fallback $\Phi_{\mathrm{llm}}$.

\subsubsection{Query-Conditioned Retrieval for the Neural Fallback}

When the symbolic solver abstains, the neural fallback retrieves and refines evidence in two steps. Let $t_q$ denote the frame associated with question $Q$. We use $h(t_q)$ to denote a textual frame anchor, and let $\mathcal{C}_q$ and $\mathcal{R}_q$ summarize the semantic categories and spatial relationship keywords expressed by the question. The retrieval backend first returns a candidate evidence set from the indexed graph view:
\begin{equation}
\mathcal{H}_Q^{0}
=
\mathcal{R}
\left(
Q \oplus h(t_q);
\mathcal{G}_{R}
\right),
\end{equation}
where $\mathcal{R}$ denotes the LightRAG retrieval
operator and $\oplus$ denotes query augmentation.

The retrieved records are subsequently refined by a deterministic query-conditioned filter:
\begin{equation}
\mathcal{H}_Q
=
\mathcal{F}_{\mathrm{filter}}
\left(
\mathcal{H}_Q^{0};
t_q,\mathcal{C}_q,\mathcal{R}_q,Q
\right).
\end{equation}
The filter removes evidence that is inconsistent with the queried frame, object semantics, or relation orientation. Its concrete implementation details are provided in Appendix~\ref{app:retrieval_filtering}.

\subsubsection{Neural Reasoning and Final Routing}

In conclusion, if the symbolic solver returns an answer, that answer is used directly. Otherwise, the filtered evidence $\mathcal{H}_Q$ is serialized into a question-type-specific prompt and passed to the LLM reasoning engine. The final prediction is
\begin{equation}
\hat{a}
=
\begin{cases}
a_{\mathrm{sym}},
&
a_{\mathrm{sym}}\neq\bot,
\\[3pt]
\Phi_{\mathrm{llm}}
\left(
Q,\mathcal{H}_{Q}
\right),
&
a_{\mathrm{sym}}=\bot.
\end{cases}
\end{equation}

This design reserves neural reasoning for cases that require semantic disambiguation, while deterministic cases are handled directly by the symbolic engine.

\section{Experiments}
\label{sec:experiments}

\subsection{Experimental Setup}
\label{subsec:experimental_setup}

\paragraph{Dataset and evaluation protocol.}
We evaluate on 5,916 NuScenes-QA questions associated with all 404 keyframes from the ten nuScenes v1.0-mini scene sequences~\cite{caesar2020nuscenes,qian2024nuscenes}. We select these questions from the union of the released training and validation question files using their sample tokens. Each question is evaluated at its associated keyframe using the corresponding scene-level graph. Per-scene and per-type statistics are provided in Appendices~\ref{app:per_scene_statistics} and~\ref{app:detailed_statistics}.

\paragraph{Oracle-perception setting.}
We use ground-truth 3D bounding boxes, object categories, instance identities, object attributes, and ego poses to construct the spatiotemporal scene graphs. This oracle-perception setting isolates graph construction, retrieval, and reasoning behavior from errors introduced by learned perception modules. Thus, the experiments measure reasoning under structured 3D observations rather than end-to-end perception accuracy.

\paragraph{LLM backends and compared configurations.}
We evaluate two LLM backends: DeepSeek-V4-Flash and GPT-5.4-mini. No task-specific model fine-tuning is performed. For the neural configurations, we keep the graph facts, embedding model, retrieval configuration, query-conditioned filtering rules, prompting strategy, and answer normalization procedure fixed across the two LLM backends.

We compare the following configurations:

\begin{itemize}
    \item \textbf{Full}: the complete framework. The selective symbolic solver first attempts to answer each question directly from the structured scene graph. When the symbolic solver abstains, the question is processed by the graph retrieval, evidence filtering, and LLM reasoning pipeline.

    \item \textbf{LLM-only}: a direct ablation of the selective symbolic route. The symbolic solver is disabled, and every question is processed by graph retrieval, query-conditioned evidence filtering, and LLM reasoning.

    \item \textbf{Symbolic-only}: a template-aware symbolic baseline. This baseline operates directly on the same unfiltered spatiotemporal scene graph and uses an expanded deterministic parser and graph-execution rule set. It attempts all five question types, without querying the graph retrieval or invoking an LLM.
\end{itemize}

The Symbolic-only configuration is intended to measure the capability of deterministic graph reasoning without neural inference. Its rule set is broader than the selective symbolic solver used in the Full framework, particularly because it contains additional rules for comparison questions. Therefore, Symbolic-only is treated as a standalone baseline rather than a strict ablation of the Full configuration. Since it does not use an LLM, its result is independent of the selected LLM backend and is reported only once.

\paragraph{Evaluation metric.}
Following NuScenes-QA~\cite{qian2024nuscenes}, we report the averaged accuracy for each question type over the complete evaluation set. Predictions are canonicalized to the corresponding answer space before evaluation. For the Symbolic-only baseline, unresolved object or status queries are marked as invalid. For the neural configurations, API failures, unparsable responses, and outputs outside the valid answer space are marked as invalid. Invalid cases remain in the evaluation denominator and are scored as incorrect.

\subsection{Main Results}
\label{subsec:main_results}

Table~\ref{tab:main_results} reports the results of the complete framework, the LLM-only ablation, and the standalone Symbolic-only baseline. The values in the table are accuracies in percent.

\begin{table*}[t]
\centering
\caption{
Answer accuracy (\%) on all ten nuScenes v1.0-mini scene sequences. The best result in each column is shown in bold. Symbolic-only is a standalone template-aware symbolic baseline and is not a strict ablation of Full.
}
\label{tab:main_results}
\setlength{\tabcolsep}{7pt}
\renewcommand{\arraystretch}{1.15}
\begin{tabular}{lcccccc}
\toprule
\textbf{Configuration}
& \textbf{Exist}
& \textbf{Count}
& \textbf{Object}
& \textbf{Status}
& \textbf{Comparison}
& \textbf{Overall} \\
\midrule
Symbolic-only
& 76.22
& 51.40
& 43.21
& 65.76
& 65.57
& 61.43 \\

LLM-only (DeepSeek)
& 82.42
& 62.56
& 59.76
& 81.63
& 66.44
& 71.25 \\

Full (DeepSeek)
& 85.52
& 75.17
& 69.32
& 89.02
& 68.67
& 77.89 \\

LLM-only (GPT)
& 84.33
& 65.45
& 61.77
& 86.30
& \textbf{76.21}
& 75.15 \\

Full (GPT)
& \textbf{88.80}
& \textbf{75.65}
& \textbf{69.88}
& \textbf{93.26}
& 74.08
& \textbf{80.63} \\
\bottomrule
\end{tabular}
\end{table*}

The Full framework with GPT-5.4-mini achieves the highest overall accuracy, reaching 80.63\% and correctly answering 4,770 of the 5,916 questions. The corresponding Full configuration with DeepSeek-V4-Flash achieves 77.89\%. Among all evaluated configurations, Full GPT obtains the best results on existence, counting, object, and status questions, while GPT LLM-only obtains the best result on comparison questions. When comparing the two Full configurations directly, GPT-5.4-mini outperforms DeepSeek-V4-Flash on all five question types.

\begin{figure}[htbp]
    \centering
    \includegraphics[width=0.49\linewidth]{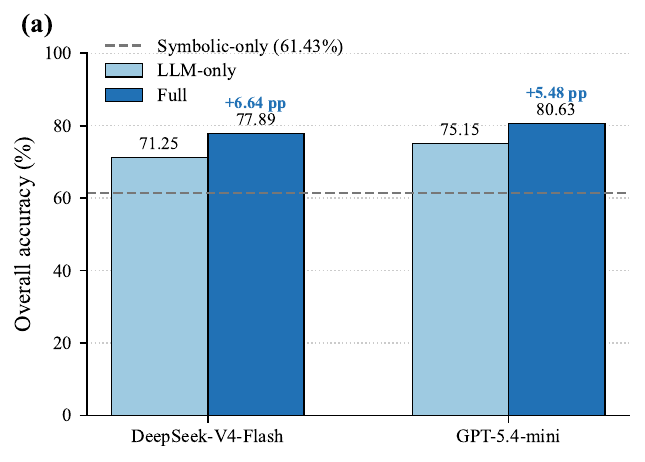}%
    \hfill
    \includegraphics[width=0.49\linewidth]{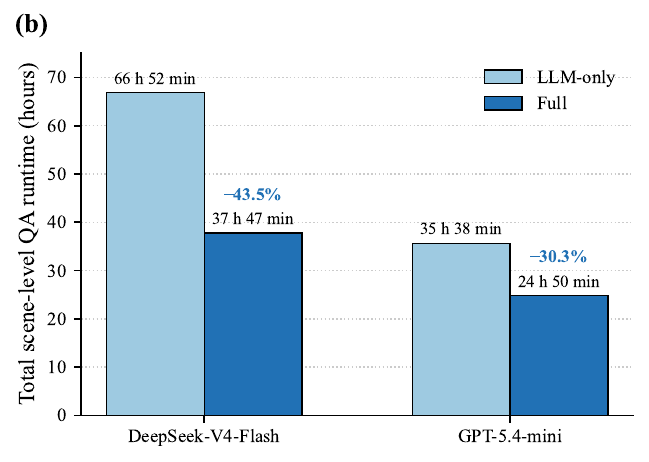}
    \caption{
    \textbf{Accuracy and scene-level QA runtime of the evaluated configurations under
    two LLM backends.}
    (a) Overall accuracy on the 5,916-question evaluation set. 
    (b) Total scene-level QA wall-clock time summed over the ten scene sequences for the LLM-only and Full configurations. The Symbolic-only baseline is omitted from this hour-scale runtime panel because it uses only local deterministic graph operations, involves no remote API calls, and completes in negligible time relative to the neural configurations.
    Blue annotations indicate accuracy gains in percentage points (pp) in (a) and relative runtime reductions in (b).
    }
    \label{fig:accuracy_runtime}
\end{figure}

\paragraph{Effect of selective symbolic execution.}
We measure the contribution of the selective symbolic route by comparing Full with LLM-only under the same LLM backend. As shown in Figure~\ref{fig:accuracy_runtime}(a), adding symbolic execution improves the overall accuracy of DeepSeek-V4-Flash from 71.25\% to 77.89\%, corresponding to a gain of 6.64 percentage points. Under GPT-5.4-mini, the overall accuracy increases from 75.15\% to 80.63\%, corresponding to a gain of 5.48 points.

The largest gains occur on counting questions. The DeepSeek-based system improves from 62.56\% to 75.17\%, a gain of 12.61 points, whereas the GPT-based system improves from 65.45\% to 75.65\%, a gain of 10.20 points. These results indicate that exact candidate-set operations and identity filtering complement neural reasoning, particularly for questions requiring enumeration or precise object selection.

\paragraph{Accuracy--efficiency trade-off.}
In addition to accuracy, we measure the wall-clock time of the scene-level QA stage. Timing begins after constructing the scene graph, and ends after all questions in the scene have been answered and parsed. The measured interval includes symbolic execution, graph retrieval, evidence filtering, neural answer generation when required, response parsing, and API retry delays. Graph construction and index preparation are excluded.

As shown in Figure~\ref{fig:accuracy_runtime}(b), across the ten scene sequences, the Full configuration reduces the total QA runtime from 66~h~52~min to 37~h~47~min with the DeepSeek backend, corresponding to a 43.5\% reduction. With GPT-5.4-mini, the runtime decreases from 35~h~38~min to 24~h~50~min, corresponding to a 30.3\% reduction. Dividing the total runtime by the 5,916 questions gives an amortized runtime per question of 40.69~s versus 22.99~s for DeepSeek, and 21.68~s versus 15.11~s for GPT-5.4-mini, for LLM-only and Full, respectively. Together with the accuracy gains in Figure~\ref{fig:accuracy_runtime}(a), these measurements show that Full improves both accuracy and runtime under the evaluated conditions. This result is consistent with the intended benefit of answering supported questions through symbolic execution before invoking retrieval and neural inference.

\paragraph{Standalone Symbolic-only baseline.}
The standalone Symbolic-only baseline achieves 61.43\% overall accuracy without querying LightRAG or invoking an LLM. It obtains 76.22\% on existence questions and 65.57\% on comparison questions, showing that deterministic template parsing combined with explicit graph execution can solve a substantial subset of the benchmark.

However, the baseline remains below both neural configurations, especially on counting and object questions. This gap reflects the limited coverage of fixed symbolic rules when questions contain linguistic variation, ambiguous references, or multiple possible candidate objects. The result motivates the combination of symbolic execution with a retrieval-augmented neural fallback.

Because Symbolic-only uses an expanded rule set, its result is not interpreted as the direct performance obtained by merely deleting the LLM from Full. Instead, it provides an independent reference for the upper range of deterministic reasoning implemented in this study.

\paragraph{Effect of the LLM backend.}
GPT-5.4-mini achieves higher overall accuracy than DeepSeek-V4-Flash in both Full and LLM-only configurations. Detailed statistics are provided in Appendix~\ref{app:detailed_statistics}.

Comparison questions require separate interpretation when comparing Full with LLM-only under the same LLM backend. In Full, the symbolic solver abstains on comparison questions, so both configurations answer these questions through the neural path. Therefore, the observed difference between Full and LLM-only on comparison questions under the same LLM backend cannot be attributed to symbolic answer execution. It may instead reflect run-to-run stochasticity or differences in neural retrieval and answer generation behavior.

\section{Discussion}
\label{sec:discussion}

The improvements across two LLM backends support selective symbolic execution as a useful strategy for structured driving-scene QA. Supported queries bypass retrieval and neural generation, while unresolved questions retain access to the neural fallback. Explicit graph facts and filtered evidence also make intermediate information inspectable. Additional discussion of these architectural properties is provided in Appendix~\ref{app:architectural_discussion}.

The evaluation is limited to frame-specific questions from ten nuScenes v1.0-mini scenes under an oracle-perception setting. It does not establish robustness to learned perception errors or the benefit of temporal links for explicit cross-frame reasoning. Symbolic answers depend on query parsing and graph facts, while the neural fallback remains sensitive to missing retrieval evidence and overly restrictive filtering. Future work will evaluate larger datasets, learned perception, and questions requiring reasoning across multiple frames.

\section{Conclusions}
\label{sec:conclusions}

We presented a neuro-symbolic question-answering framework for autonomous driving scenarios that combines an explicit spatiotemporal scene graph with selective symbolic execution and retrieval-augmented language reasoning. The hierarchical graph separates persistent object identities from frame-specific states and explicitly represents spatial relations and temporal transitions. The symbolic solver answers unambiguous queries through deterministic graph operations and delegates unresolved questions to an LLM.

Experiments on NuScenes-QA questions across all ten nuScenes v1.0-mini scenes demonstrate improvements in both accuracy and QA efficiency under an oracle-perception setting. These results support selective symbolic execution as an effective way to improve answer accuracy while reducing reliance on retrieval and neural inference.

The current evaluation focuses on frame-specific questions over ground-truth scene observations. Future work will integrate learned perception, evaluate robustness to incomplete or noisy graph evidence, and extend the evaluation to larger datasets and questions that explicitly require reasoning across multiple frames. These extensions will help assess the framework's scalability and the contribution of its temporal structure to more demanding driving-scene reasoning tasks.

\section*{AI Assistance}
Large language models were used to assist with coding, language polishing, drafting manuscript sections, suggesting relevant literature, and discussing methodological choices. The authors take full responsibility for the manuscript, including the accuracy of its references, methods, results, and conclusions. The use of LLMs as experimental reasoning backends is described in the methodology and experimental setup.



\bibliography{iclr2027_conference}
\bibliographystyle{iclr2027_conference}

\newpage
\appendix

\section{Appendix: Additional Implementation and Details}
\label{app:additional_details}

\subsection{Per-Scene Evaluation Statistics}
\label{app:per_scene_statistics}

Table~\ref{tab:scene_statistics} lists the ten scene sequences used in the evaluation. The number of frames is obtained by following the sample chain from each scene's first sample to its last sample. The question count includes all retained NuScenes-QA questions whose sample tokens belong to the corresponding scene sequence.

\begin{table*}[t]
\centering
\caption{
Per-scene frame and question statistics for the nuScenes v1.0-mini
evaluation subset.
}
\label{tab:scene_statistics}
\setlength{\tabcolsep}{8pt}
\renewcommand{\arraystretch}{1.1}
\begin{tabular}{c c c c}
\toprule
\textbf{Scene index}
& \textbf{Scene name}
& \textbf{Frames}
& \textbf{Questions} \\
\midrule
0 & scene-0061 & 39 & 645 \\
1 & scene-0103 & 40 & 478 \\
2 & scene-0655 & 41 & 550 \\
3 & scene-0553 & 41 & 751 \\
4 & scene-0757 & 41 & 842 \\
5 & scene-0796 & 40 & 695 \\
6 & scene-0916 & 41 & 608 \\
7 & scene-1077 & 41 & 335 \\
8 & scene-1094 & 40 & 584 \\
9 & scene-1100 & 40 & 428 \\
\midrule
\textbf{Total}
& \textbf{10 sequences}
& \textbf{404}
& \textbf{5,916} \\
\bottomrule
\end{tabular}
\end{table*}

The sequence length limit is set to 41, which covers the longest scene in the v1.0-mini release. Scenes with fewer than 41 keyframes use all of their available keyframes. In the present evaluation, the ten sequences contain 39--41 keyframes, so no scene requires padding or truncation. Questions are selected using their original sample tokens, ensuring that each question remains associated with its original keyframe.

\subsection{Ground-Truth Spatiotemporal Graph Construction}
\label{app:gt_graph_construction}

For every keyframe, we read the ground-truth object annotations from nuScenes. Each annotation provides an object category, an instance identity, an attribute description, and a 3D bounding box. The bounding box is transformed into the ego-vehicle coordinate system using the ego pose associated with the front camera. The transformed BEV center is used to determine the ego-centric spatial relation.

A persistent entity is created for each instance identity appearing in the scene sequence. For every observed instance at frame $t$, the graph contains a corresponding temporal state node with a unique frame-specific identifier. The state node stores the frame association, object category, motion status, and other attributes required by the downstream reasoning modules.

The graph contains the following edge types:

\begin{itemize}
    \item \textbf{Ego-centric spatial edges}: each state node is connected to the ego-reference node using one of the six directional predicates defined by NuScenes-QA;

    \item \textbf{Object-to-object spatial edges}: directed spatial edges are generated between ordered pairs of objects observed at the same frame;

    \item \textbf{Hierarchy edges}: a persistent entity is connected to each of its frame-specific state nodes through a \texttt{HAS\_STATE} edge;

    \item \textbf{Temporal edges}: successive observations of the same persistent instance are connected through a \texttt{NEXT\_STATE} edge.
\end{itemize}

\begin{figure}[H]
    \centering
    \includegraphics[width=\textwidth]{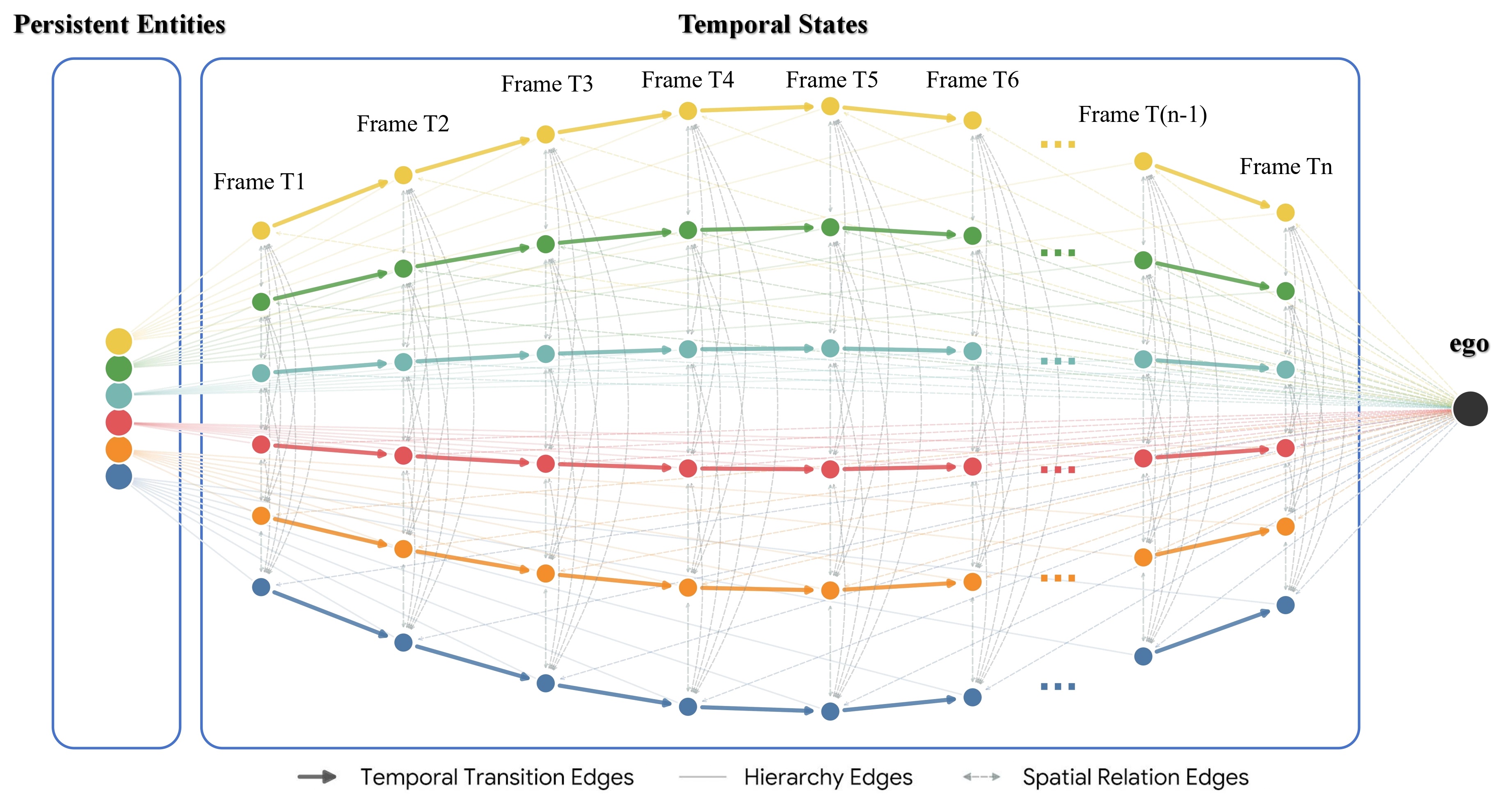}
    \vspace{-6mm} 
    \caption{\textbf{Hierarchical spatiotemporal scene graph.} Each color denotes a persistent entity and its frame-specific states. Hierarchy edges link entities to their states, thick arrows connect successive observations of the same entity, and dashed edges represent directed spatial relations between states within each frame and between states and the ego vehicle.}
    \label{fig:Graph}
\end{figure}

The spatial predicate is computed from the subject-relative displacement between two object centers in the ego coordinate system. We use the six-direction taxonomy and angular convention defined by NuScenes-QA~\cite{qian2024nuscenes}; the exact angular intervals are therefore not repeated here.

Object-to-object spatial relations are generated without an explicit distance threshold. To avoid uninformative background-background connections, relations whose two endpoints are both traffic cones or barriers are excluded. The graph therefore retains long-range object relations while controlling redundant relations between static traffic-control objects.

For motion status, the implementation uses the following category-aware defaults when no explicit attribute is available: vehicles are initialized as \texttt{stopped}, pedestrians as \texttt{standing}, bicycles and motorcycles as \texttt{withoutrider}, and barriers and traffic cones as \texttt{none}. When an attribute is present, the corresponding attribute-derived status overrides the default.

\subsection{Graph Projection and LightRAG Configuration}
\label{app:lightrag_configuration}

The complete graph is saved as a scene-level JSON object and remains available to the symbolic reasoning module. For retrieval, the graph is projected into a task-oriented structured representation and inserted through LightRAG's custom-KG interface~\cite{guo2025lightrag}. The main indexed view contains a scene-summary record, frame-specific state records, and spatial-relation records.

Each indexed relation is represented by its source identifier, target identifier, directional predicate, and frame association. Its canonical description preserves the subject--predicate--reference ordering, for example: \texttt{Fact: A IS FRONT\_OF B. frame\_idx=5}. This representation keeps the direction of each spatial fact explicit.

In the reported main implementation, the persistent entity layer, \texttt{HAS\_STATE} edges, and \texttt{NEXT\_STATE} edges remain in the complete scene graph used by symbolic reasoning. The retrieval projection focuses on the records required by the current frame-specific QA tasks. This separation allows the graph to retain longer-term structural information without requiring every graph element to be retrieved for every question.

The graph is inserted directly as structured custom knowledge. The implementation does not call the ordinary document insertion path for the scene text and does not ask an LLM to re-extract the ground-truth entities and relations. The embedding model is
\texttt{text-embedding-3-small} with embedding dimension 1536. Each scene is assigned an independent LightRAG working directory.

The neural configurations use LightRAG local retrieval, and the retrieval query consists of the natural language question together with a textual anchor for the question's sample-specific frame. The main retrieval settings are summarized in Table \ref{tab:implementation_settings}.

\begin{table}[t]
\centering
\caption{
Main retrieval and indexing settings.
}
\label{tab:implementation_settings}
\setlength{\tabcolsep}{5pt}
\renewcommand{\arraystretch}{1.1}
\begin{tabular}{ll}
\toprule
\textbf{Component} & \textbf{Setting} \\
\midrule
Retrieval mode & local \\
Reranking & disabled \\
Embedding model & text-embedding-3-small \\
Embedding dimension & 1536 \\
Chunk token size & 8192 \\
Embedding batch size & 48 \\
Entity extraction gleaning & 0 \\
\bottomrule
\end{tabular}
\end{table}

\subsection{Query-Conditioned Retrieval and Evidence Filtering}
\label{app:retrieval_filtering}

The target frame is obtained from the question's sample token through the scene-level token-to-frame mapping. It is not inferred from the natural language question. For questions that reach the neural fallback, the system executes the following two-step process:

\begin{equation}
\mathcal{H}_Q^0
=
\mathcal{R}_{\mathrm{LR}}
\left(
Q \oplus h(t_q);
\mathcal{G}_R
\right),
\end{equation}

\begin{equation}
\mathcal{H}_Q
=
\mathcal{F}_{\mathrm{filter}}
\left(
\mathcal{H}_Q^0;
t_q,\mathcal{C}_q,\mathcal{R}_q,Q
\right).
\end{equation}

First, LightRAG retrieves candidate entities and relations using the question and the frame anchor. Second, the returned serialized evidence is filtered using deterministic compatibility rules. The filtering procedure applies four practical constraints:

\begin{enumerate}
    \item \textbf{Frame constraint}: records associated with frames other than the queried frame are removed when an exact frame is available;

    \item \textbf{Semantic constraint}: records belonging to irrelevant object categories are removed. For queries containing generic terms such as \textit{thing} or \textit{object}, we relax category filtering to retain retrieved evidence across traffic-participant categories. Barriers and traffic cones are retained only when explicitly mentioned in the query, either as targets or reference objects;

    \item \textbf{Directional constraint}: spatial relations are retained only when their directional predicate matches a spatial expression in the question. If a question contains no spatial constraint, spatial relation records are removed;

    \item \textbf{Ego-reference constraint}: ego-relative relation records are retained only when the question refers to the ego vehicle through expressions such as \textit{me}, \textit{ego}, or \textit{my car}.
\end{enumerate}

The filtering step operates on the serialized retrieval result before it is inserted into the LLM prompt. The resulting evidence is passed to a question-type specific prompt.

\subsection{Symbolic Routing and Baseline Details}
\label{app:symbolic_details}

The Full configuration uses a conservative symbolic-first route. The symbolic solver directly reads the complete Python scene graph and attempts to resolve the question before LightRAG retrieval is performed. Its behavior can be summarized as:

\begin{equation}
\hat{a}
=
\begin{cases}
\Phi_{\mathrm{sym}}(Q,\mathcal{G},t_q),
&
\Phi_{\mathrm{sym}}(Q,\mathcal{G},t_q)
\neq \bot,
\\[3pt]
\Phi_{\mathrm{llm}}(Q,\mathcal{H}_Q),
&
\Phi_{\mathrm{sym}}(Q,\mathcal{G},t_q)
=
\bot.
\end{cases}
\end{equation}

The symbolic solver supports deterministic existence and counting queries, as well as object and status queries when the remaining candidates yield a unique answer. It counts persistent global identifiers rather than frame-specific state identifiers. Comparison questions and unsupported multi-hop patterns are delegated to the neural fallback in the Full configuration.

The Symbolic-only baseline uses a different, expanded solver. It directly loads the unfiltered scene graph saved by the main experiment and does not query LightRAG. Its parser recognizes object categories, motion statuses, distance bands, ego-relative spatial expressions, and nested spatial descriptions. It also contains comparison-specific rules, which allows it to attempt comparison questions without an LLM.

The Symbolic-only baseline applies the following answer conditions:

\begin{itemize}
    \item existence questions return \texttt{yes} or \texttt{no} according to whether the constrained candidate set is empty;

    \item counting questions count unique persistent entity identifiers;

    \item object questions return a category only when the candidate categories are unique;

    \item status questions return a status only when one persistent entity and one status are identified;

    \item comparison questions resolve two descriptors and compare the selected attribute, such as motion status, category, direction, distance band, or color.
\end{itemize}

When an object or status answer cannot be uniquely determined, the Symbolic-only implementation returns an invalid result. For comparison questions, unresolved or non-unique descriptors are handled by the deterministic comparison rule used by the baseline. These behaviors are included to make the reported baseline reproducible.

\subsection{Prompting and Answer Normalization}
\label{app:prompting_normalization}

The neural fallback uses question-type-specific prompts for existence, counting, object, status, and comparison questions. The prompt contains the original question, the filtered evidence, and the corresponding output constraints. The required final format is a JSON object of the form \texttt{\{"answer": "value"\}}.

The answer spaces are:

\begin{itemize}
    \item \textbf{Existence}: \texttt{yes} or \texttt{no};

    \item \textbf{Counting}: a non-negative integer;

    \item \textbf{Object}: one of the supported object categories;

    \item \textbf{Status}: one of the supported motion-status labels;

    \item \textbf{Comparison}: \texttt{yes} or \texttt{no}.
\end{itemize}

Before scoring, capitalization, punctuation, spaces, underscores, and known aliases are normalized. For example, \textit{traffic cone} is mapped to \texttt{trafficcone}, \textit{construction vehicle} is mapped to \texttt{constructionvehicle}, and \textit{person} is mapped to \texttt{pedestrian}. Object names with trailing instance numbers are reduced to their category.

The implementation also applies category-aware status conventions. For example, \texttt{withrider} and \texttt{withoutrider} are used for bicycles and motorcycles, while \texttt{moving}, \texttt{parked}, and \texttt{stopped} are used for vehicles. These conventions prevent status labels from being assigned across incompatible object categories.

\subsection{Detailed Prediction Statistics}
\label{app:detailed_statistics}

Table~\ref{tab:detailed_statistics} reports the number of correct and invalid predictions for every configuration. Each cell is formatted as \textit{correct/invalid}. Valid but incorrect predictions are not included in the invalid count.

\begin{table*}[t]
\centering
\caption{
Detailed prediction statistics on the complete v1.0-mini evaluation subset. Each entry reports \textit{correct/invalid}; invalid cases remain in the denominator and are scored as incorrect.
}
\label{tab:detailed_statistics}
\setlength{\tabcolsep}{3.5pt}
\renewcommand{\arraystretch}{1.15}
\resizebox{\textwidth}{!}{%
\begin{tabular}{l r c c c c c}
\toprule
\textbf{Question type}
& \textbf{Total}
& \textbf{Symbolic-only}
& \textbf{LLM-only (DeepSeek)}
& \textbf{Full (DeepSeek)}
& \textbf{LLM-only (GPT)}
& \textbf{Full (GPT)} \\
\midrule
Exist
& 1,678
& 1,279/0
& 1,383/9
& 1,435/1
& 1,415/6
& 1,490/9 \\

Count
& 1,039
& 534/0
& 650/35
& 781/20
& 680/0
& 786/6 \\

Object
& 1,245
& 538/518
& 744/394
& 863/295
& 769/3
& 870/17 \\

Status
& 920
& 605/253
& 751/4
& 819/2
& 794/0
& 858/3 \\

Comparison
& 1,034
& 678/0
& 687/54
& 710/32
& 788/0
& 766/17 \\

\midrule
Overall
& 5,916
& 3,634/771
& 4,215/496
& 4,608/350
& 4,446/9
& 4,770/52 \\
\bottomrule
\end{tabular}%
}
\end{table*}

\subsection{Scene-level Runtime Measurement}
\label{app:runtime_measurement}

We additionally measure the wall-clock time required to answer all questions associated with each scene sequence. The evaluation is performed sequentially in the original question order. For each scene, the timer starts after the scene graph and the prebuilt LightRAG index have been loaded and initialized, and stops after the last question has been answered and parsed. The measured interval includes symbolic execution, LightRAG retrieval, query-conditioned evidence filtering, neural answer generation, response parsing, network waiting time, and API retry delays.

The timer excludes nuScenes database initialization, scene-graph construction, graph indexing, embedding computation during index construction, API preflight checks, and final result-file serialization. Therefore, the reported quantity is scene-level QA inference time with prebuilt graph and retrieval indexes rather than end-to-end perception or simulator execution time.

For each question, we also record the individual wall-clock latency and the execution route. A question is assigned to the symbolic route when the solver returns a valid answer; otherwise, it is assigned to the neural fallback route. Failed and invalid queries remain included in the total runtime and in the accuracy denominator.

The ten scenes contain 5,916 questions in total and are processed without inter-question parallelism. Consequently, the reported scene-level time reflects sequential evaluation throughput. Because the neural path includes remote API calls, the absolute runtime depends on backend serving latency, network conditions, queueing, and retry behavior. The paired Full-versus-LLM-only comparison under the same backend is therefore the primary comparison for measuring the efficiency contribution of selective symbolic execution.

\begin{table*}[t]
\centering
\caption{
Scene-level QA wall-clock runtime for the ten nuScenes v1.0-mini scene sequences. Runtime is reported as hours:minutes (h:mm). Reduction denotes the relative runtime decrease of Full compared with LLM-only under the same LLM backend.
}
\label{tab:scene_runtime}

\small
\setlength{\tabcolsep}{6pt}
\renewcommand{\arraystretch}{1.1}

\begin{tabular}{l r r r r r r}
\toprule
& \multicolumn{3}{c}{\textbf{GPT}}
& \multicolumn{3}{c}{\textbf{DeepSeek}} \\
\cmidrule(lr){2-4}
\cmidrule(lr){5-7}
\textbf{Scene}
& \textbf{Full}
& \textbf{LLM-only}
& \textbf{Reduction}
& \textbf{Full}
& \textbf{LLM-only}
& \textbf{Reduction} \\
\midrule
scene0 & 4:36 & 8:08 & 43.4\% & 8:03 & 12:49 & 37.2\% \\
scene1 & 2:45 & 2:54 & 5.2\%  & 3:05 & 4:59  & 38.1\% \\
scene2 & 3:22 & 4:19 & 22.0\% & 2:41 & 6:38  & 59.5\% \\
scene3 & 3:35 & 4:24 & 18.6\% & 5:15 & 10:12 & 48.5\% \\
scene4 & 1:22 & 2:20 & 41.4\% & 4:11 & 8:09  & 48.7\% \\
scene5 & 1:35 & 2:02 & 22.1\% & 3:12 & 5:11  & 38.3\% \\
scene6 & 3:25 & 4:49 & 29.1\% & 4:24 & 6:50  & 35.6\% \\
scene7 & 0:59 & 1:02 & 4.8\%  & 1:08 & 2:07  & 46.5\% \\
scene8 & 2:32 & 4:19 & 41.3\% & 3:57 & 5:35  & 29.3\% \\
scene9 & 0:39 & 1:21 & 51.9\% & 1:51 & 4:22  & 57.6\% \\
\midrule
\textbf{Total}
& \textbf{24:50}
& \textbf{35:38}
& \textbf{30.3\%}
& \textbf{37:47}
& \textbf{66:52}
& \textbf{43.5\%} \\
\bottomrule
\end{tabular}
\end{table*}

Across all ten scenes, GPT-5.4-mini requires 24~h~50~min in the Full configuration and 35~h~38~min in the LLM-only configuration. DeepSeek-V4-Flash requires 37~h~47~min and 66~h~52~min, respectively. Thus, the Full configuration saves 10~h~48~min with GPT-5.4-mini and 29~h~05~min with DeepSeek-V4-Flash. The reduction is observed in every scene for both backends, although its magnitude varies with the retrieval cost and backend serving latency.

\subsection{Additional Discussion of Architectural Properties}
\label{app:architectural_discussion}

Driving vision-language models such as DriveLM use neural inference for driving-scene question answering~\cite{sima2025DriveLM}, while VISPROG uses an LLM to generate executable visual programs~\cite{Gupta2023Visual}. Graph-based RAG frameworks such as LightRAG retrieve structured evidence for neural answer generation~\cite{guo2025lightrag}. Our framework places a conservative symbolic executor before both retrieval and LLM inference, allowing supported, unambiguous queries to be answered directly from the scene graph. This selective design reduces reliance on neural computation for supported queries and provides explicit graph evidence for inspecting the answering process. Its main architectural properties are discussed below:


\begin{enumerate}
    \item \textbf{Deterministic Graph Operations with Selective Execution:} For supported queries, the symbolic solver evaluates explicit spatial relations, filters candidate objects, and performs set operations over persistent identities. This provides deterministic results for the parsed constraints on the available graph. Answer correctness nevertheless depends on the graph facts and the interpretation of the question. The solver therefore abstains on unsupported patterns or unresolved ambiguities and delegates these cases to the neural fallback.

    \item \textbf{Reasoning without Task-Specific Fine-Tuning:} The framework uses pretrained LLMs without task-specific parameter updates. Task adaptation is implemented through the graph schema, symbolic rules, evidence filtering, and question-type-specific prompts. The experiments demonstrate this approach with two LLM backends.

    \item \textbf{Inspectable Evidence and Intermediate Results:} Explicit entity identifiers, frame associations, and directed relations make the evidence used for answering questions inspectable. Symbolic candidate sets and filtered retrieval records provide intermediate results that can support error analysis.

    \item \textbf{Explicit Identity and Temporal Structure:} Persistent entity nodes connect observations of the same object across frames, while \texttt{NEXT\_STATE} edges record successive observed states. This representation organizes scene information across a sequence and provides a basis for temporal queries. The current evaluation focuses on frame-specific questions, so the benefit of temporal transitions for cross-frame reasoning has not yet been established.

    \item \textbf{Modular Perception and Reasoning Interfaces:} Structured object observations provide an explicit interface between perception, graph construction, and reasoning. The reasoning backend can therefore be changed while retaining the graph representation and retrieval pipeline, as illustrated by the two evaluated LLM backends. This modularity does not prevent perception errors from affecting downstream answers; its robustness with learned perception requires further study.
\end{enumerate}

\subsubsection{Accuracy and QA Efficiency}
\label{subsec:discussion_efficiency}

The observed accuracy gains are accompanied by reductions in QA wall-clock time of 30.3\% with GPT-5.4-mini and 43.5\% with DeepSeek-V4-Flash relative to their corresponding LLM-only configurations. These results are consistent with the routing design: questions answered symbolically bypass retrieval and neural generation, while unresolved questions retain access to the LLM fallback. The measurements cover only the QA stage and exclude graph construction and index preparation. They therefore demonstrate improved QA efficiency under the evaluated conditions.

\subsubsection{Limitations and Future Work}
\label{subsec:discussion_limitations}

The evaluation uses ground-truth observations from ten nuScenes v1.0-mini scenes. Future work will integrate an upstream perception module through the structured observation interface to evaluate the framework on practical driving tasks. We will also introduce questions that explicitly require cross-frame reasoning to assess the effectiveness of the hierarchical structure in handling more complex reasoning tasks.

In addition, both reasoning routes retain potential sources of error. Symbolic execution depends on the coverage and correctness of the query parser. For the neural fallback, deterministic filtering can remove incompatible records but cannot recover relevant facts absent from the retrieved candidate set. Overly restrictive filters may also discard useful evidence.

\end{document}